\documentclass{article}
\usepackage{spconf,amsmath,graphicx,hyperref}
\usepackage{enumitem}
\usepackage{listings}
\usepackage{multirow}
\usepackage{float}
\usepackage{booktabs}
\usepackage{amsfonts} 

\usepackage{comment}
\usepackage[most]{tcolorbox}
\usepackage{makecell}

\newtcolorbox{promptbox}[1][]{
  colback=pink!10!white,        
  colframe=pink!50!purple,      
  fonttitle=\bfseries\scriptsize,
  title=Prompt Template,
  boxrule=0.5pt,
  arc=3mm,                      
  auto outer arc,
  coltitle=black,
  left=3mm, right=3mm, top=1mm, bottom=1mm,
  #1
}

\title{FLAME: Empowering Frozen LLMs for Knowledge Graph Completion}

\name{Bo Xue$^{\dagger}$ \quad Yi Xu$^{\dagger}$ \quad Bolei Ma$^{\ddagger}$ \quad Yunchong Song$^{\dagger}$ \quad Jiaxin Ding$^{\dagger}$\sthanks{Corresponding Author. This work was supported by NSF China under Grant No. 62525209, 92579104, T2421002, T2542021.} \quad Luoyi Fu$^{\dagger}$ \quad Xinbing Wang$^{\dagger}$}
  
  \address{$^{\dagger}$ School of Computer Science, Shanghai Jiaotong University, China \\
      $^{\ddagger}$ LMU Munich \& Munich Center for Machine Learning, Germany}

\begin{document}
\ninept
\maketitle
\begin{abstract}
Traditional knowledge graph completion (KGC) methods rely solely on structural information and struggle with sparsity, while Large Language Models (LLMs) address these limitations through rich world knowledge and strong context modeling. Fine-tuning LLMs is effective but costly, while non-fine-tuned LLMs are efficient but suboptimal. To address this trade-off, we propose \textbf{FLAME}, a framework that extracts context-aware hidden states from intermediate layers of frozen LLMs to train data-efficient KGC classifiers. We bridge LLM-KG semantic gaps via subgraph-based entity descriptions and employ sliced mutual information (SMI) to quantify task-relevant information in representations. Experiments demonstrate that FLAME achieves 47\% improvement over non-fine-tuned LLM baselines and, to our knowledge, is the first to achieve fine-tuned performance with $188\times$ memory efficiency and $26.11\times$ speedup.

\end{abstract}
\begin{keywords}
Knowledge Graph Completion, Large Language Models, Data-Efficient Learning
\end{keywords}
\section{Introduction}
\label{sec:intro}
\begin{figure*}[t]
    \centering
    \includegraphics[width=0.85\linewidth]{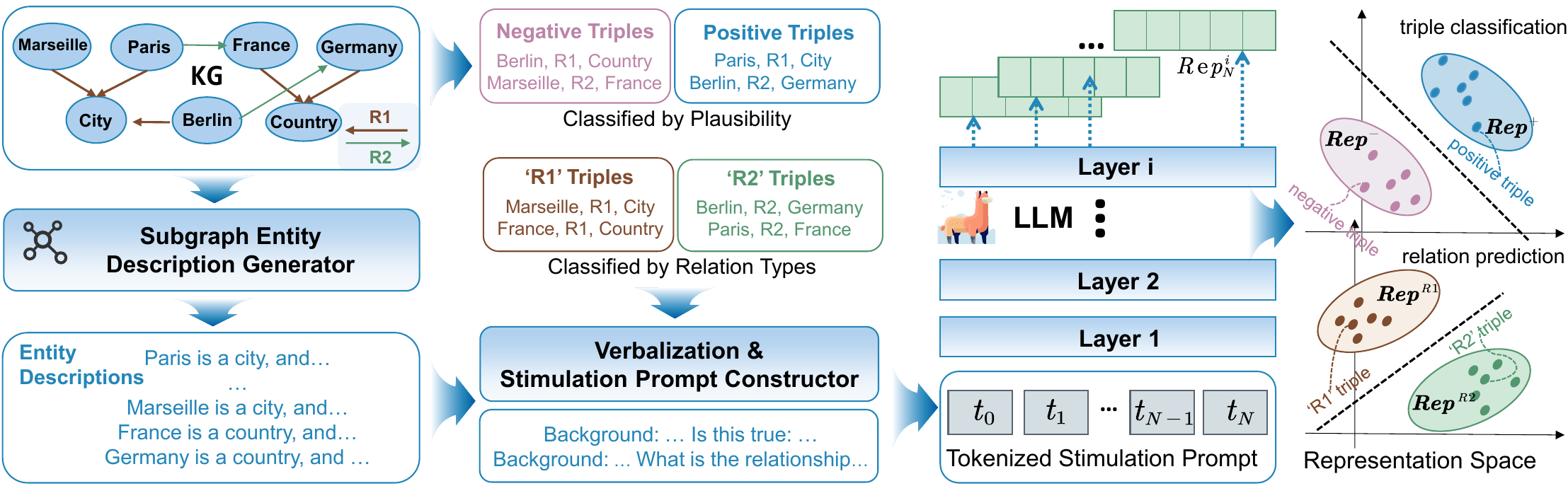}
    \caption{Overview of \textbf{FLAME}. Triples are categorized by task and verbalized, then combined with subgraph-derived entity descriptions to form stimulation prompts. Context-aware hidden states are extracted from a frozen LLM and used to train a data-efficient KGC classifier. \textbf{Only the classifier requires training},  enabling low memory usage, fast training and inference.}

    \label{fig:arc}
    \vspace{-2mm}
\end{figure*}

Knowledge graph completion aims to expand knowledge graphs by predicting potential relationships and discovering new facts \cite{chen2020knowledge, shen2022comprehensive}. Traditional knowledge graph embedding methods rely solely on structural information \cite{bordes2013translating,trouillon2016complex,sun2019rotate} and are vulnerable to graph sparsity \cite{yao2019kg}. Recent work shows that LLMs can mitigate these limitations through contextual modeling and implicit world knowledge~\cite{wei2024kicgpt}.

Existing LLM-based KGC approaches fall into two categories: non-fine-tuned LLMs with prompt techniques, which are computationally efficient but may underperform due to task-specific misalignment \cite{wei2024kicgpt,yang2024enhancing}, and fine-tuned LLMs, which improve performance through better task alignment but require substantial computational resources \cite{yao2025exploring, zhang2024making}. This raises the key question: \emph{How can we leverage LLMs for effective KGC without the cost of fine-tuning?}

To address this challenge, we propose FLAME, which achieves the benefits of fine-tuning—reduced hallucinations and improved semantic alignment—without parameter updates~\cite{zhou2024lima}. FLAME extracts context-aware hidden states from frozen LLMs to train a lightweight classifier (Fig.~\ref{fig:arc}). It samples subgraphs to generate model-friendly entity descriptions that bridge LLM-KG semantic gap. To understand why frozen LLMs can be effective, we employ SMI from an information-theoretic perspective to quantify the task-relevant information preserved in intermediate representations. Through this approach, FLAME attains performance comparable to fine-tuned models while maintaining the efficiency of frozen LLMs.

To systematically evaluate our framework, we investigate three research questions:
\begin{itemize}[leftmargin=*,nolistsep]
\item \textbf{RQ1:} \emph{Can LLMs with our method effectively achieve performance on par with their fine-tuned counterparts?} (§\ref{sec:rq1})
\item \textbf{RQ2:} \emph{How does our method perform across different language models and classification models?} (§\ref{sec:rq2})
\item \textbf{RQ3:} \emph{What are the data efficiency and computation efficiency of our proposed method for KGC?} (§\ref{sec:rq3})
\end{itemize}

Extensive experiments demonstrate that FLAME achieves a 47\% relative improvement over existing non-fine-tuned methods, matching fine-tuned LLM performance while enhancing GPU memory efficiency by $188\times$ during training and accelerating training and inference by $26.11\times$. Our contributions are threefold:

\begin{itemize}[leftmargin=*,nolistsep]
\item We introduce FLAME, the first framework that systematically exploits intermediate representation space of frozen LLMs for KGC, enriched with a subgraph-based entity description generator to align the KG's symbolic space with the LLM's semantic space.
\item  We theoretically quantify task-specific information encoded in LLM representations using SMI and develop targeted optimization strategies that enable frozen LLMs to achieve performance parity with fine-tuned models for the first time in KGC.
\item We conduct comprehensive experiments on standard KGC benchmarks, systematically demonstrating the effectiveness and efficiency of FLAME across diverse settings.
\end{itemize}

\section{METHOD}\label{sec3}

\subsection{Task Formulation and Notations}\label{sec:formulation}

A knowledge graph $\mathcal{G}$ consists of entities $\mathcal{E}$ and relations $\mathcal{R}$, represented as triples $\mathcal{G} = \{(h, r, t)\}$ with $h, t \in \mathcal{E}$ and $r \in \mathcal{R}$. We collectively refer to the head entity $h$ and tail entity $t$ as $e$. Let $D(e)$ and $D(r)$ denote long textual descriptions of entities and relations, and $EN(e)$ the entity name. 

We consider three standard KGC settings: (1) \textbf{triple classification} — verify a triple's validity; (2) \textbf{relation prediction} — predict a missing relation; (3) \textbf{entity prediction} — predict a missing entity.

\subsection{Extracting Knowledge Representations}\label{sec:3.2}

While frozen LLMs encode accurate relational knowledge, direct prompting yields hallucinations~\cite{zou2023representation, orgad2024llms}. Fine-tuning mitigates this by optimizing parameters to produce KG-consistent outputs but is computationally costly~\cite{zhou2024lima}. Instead, we perform KGC-targeted extraction of intermediate representations as an efficient alternative to fine-tuning, exploiting frozen LLMs' internal semantics to avoid both computational overhead and generation-based hallucinations.

\begin{figure}[htb]
\centering
\begin{promptbox}[width=0.95\linewidth]
\scriptsize 
\textbf{$PT_1$:} Is this true: $Verbalize(\{EN(h)\} \{EN(r)\} \{EN(t)\})$?\\[0.5em]
\textbf{$PT_2$:} Background: 1. $\{D(h)\}$ 2. $\{D(t)\}$\\
Question: Is this true: $Verbalize(\{EN(h)\} \{EN(r)\} \{EN(t)\}$)?\\[0.5em]
\textbf{$PT_3$:} What is the relationship between $\{EN(h)\}$ and $\{EN(t)\}$?\\
Please choose your answer from: $\{r_1\}$ \textbar $\{r_2\}$ \textbar \dots \textbar $\{r_n\}$.\\[0.5em]
\textbf{$PT_4$:} Background: 1. $\{D(h)\}$ 2. $\{D(t)\}$\\
Question: What is the relationship between $\{EN(h)\}$ and $\{EN(t)\}$?\\
Please choose your answer from: $\{r_1\}$ \textbar $\{r_2\}$ \textbar \dots \textbar $\{r_n\}$.
\end{promptbox}
\vspace{-1mm}
\caption{Prompt templates to stimulate the LLM.}
\label{fig:prompt}
\vspace{-1mm}
\end{figure}

\textbf{Prompt construction.} For triple classification, we design task-specific prompts (see $PT_1$ in Fig.~\ref{fig:prompt}) to strategically stimulate plausibility-relevant representations in intermediate layers. Positive samples $S^+$ are drawn from the KG, while negative samples $S^-$ are generated via standard negative sampling~\cite{yao2019kg}.

\textbf{Representation extraction.} For each sample $s \in S^+ \cup S^-$, we extract intermediate-layer hidden states:

\begin{equation}
Rep_j^l(s) = M_{1:l}(s)[j], 0\textless l\textless L, 0\textless j\textless N
\end{equation}
where $M$ represents the language model, $M_{1:l}$ denotes the first $l$ layers and $j$ the token index, $L$ and $N$ denote the number of model layers and tokens, respectively. Following~\cite{zou2023representation}, we extract the last token’s representation for both positive and negative samples, due to the autoregressive nature of causal language models.

\textbf{Classifier training.} We train a lightweight classifier (e.g., MLP, logistic regression) using representations extracted from positive and negative samples to distinguish plausible triples. For relation prediction, we adapt this binary framework to multi-class classification using $PT_3$ in Fig.~\ref{fig:prompt} to stimulate relation-specific features.

\subsection{Subgraph Entity Description Generator}\label{sec:3.3}

Semantic ambiguity in KG entities poses significant challenges for frozen LLMs relying solely on entity names. Fine-tuning alleviates this by aligning LLMs' internal entity understanding with KG entities but is computationally expensive. We propose a subgraph-based approach that derives entity descriptions from local subgraph structures, enabling comparable semantic alignment for frozen LLMs.

Since most KGs lack entity descriptions $D(e)$, we propose two generation approaches: \textbf{(1) Structured verbalization:} Unlike \cite{yang2024enhancing}'s single-triple approach, we leverage richer subgraph contexts. Following \cite{zhang2024making}, $D_1(e)$ concatenates verbalized one-hop triples:  
\begin{equation}
    \begin{aligned}
        D_1(e)=\operatorname{CONCAT}\big(\{V(EN(h_i),EN(r_i),EN(t_i))|\\
        (h_i,r_i,t_i)\in \operatorname{Subgraph}(e)\}\big),
    \end{aligned}
\end{equation}
where $\operatorname{Subgraph(\cdot)}$ retrieves one-hop triples, $\operatorname{V(\cdot)}$ denotes linear verbalization \cite{baek2023knowledge}, and $\operatorname{CONCAT(\cdot)}$ concatenates with separators.  \textbf{(2) Model-friendly narrative:} Since structured triples deviate from LLM pretraining distributions \cite{bian2024rulestorybettercommonsense}, we employ in-context learning (Fig.~\ref{fig:des_prompt}) to transform $D_1(e)$ into a model-friendly narrative $D_2(e)$ that preserves local subgraph constraints, improving knowledge retrieval during KGC.  
\begin{figure}[th!]
\centering
\begin{promptbox}[width=0.95\linewidth]
\scriptsize 
\textbf{User:} The entity description is a description of the entity name. Given the entity name: `$\{EN(e)\}$'; at the same time, `$\{EN(e)\}$' satisfies the constraints: $\{D_1(e)\}$. Please generate an entity description that satisfies the constraints for `$\{EN(e)\}$'.

\textbf{Assistants:}
\end{promptbox}
\vspace{-1mm}
\caption{Prompt template for generating $D_2(e)$.}
\vspace{-1mm}
\label{fig:des_prompt}
\end{figure}

These descriptions form a principled bridge from the KG's symbolic space to the LLM's semantic space, providing structurally-aware narratives that encode local graph constraints.  The alignment akin to entity linking underpins prompts $PT_2$ and $PT_4$ (Fig.~\ref{fig:prompt}).

\subsection{Representation Evaluation via SMI}\label{sec:3.4}

To validate that frozen LLM representations preserve sufficient task-relevant information for effective KGC, we employ SMI, which scales more effectively to high-dimensional representations than mutual information~\cite{goldfeld2021sliced, wongso2023using}. Let the random variable $X$ denote representations from intermediate layers, where $x_k = Rep_j^l(s_k)\in\mathbb{R}^d$ for sample $k$, and $Y$ the discrete KGC labels. SMI is defined as the expectation of mutual information between $Y$ and random one-dimensional projections of $X$, sampled uniformly from the $d$-dimensional unit sphere $\mathbb{S}^{d-1}$. It is estimated via Monte-Carlo sampling with a scalar mutual information estimator~\cite{kraskov2004estimating}:

\begin{equation}
    \begin{aligned}
        \widehat{SMI}(X;Y)=\frac{1}{m}\sum_{i=1}^{m}
        \hat{I}\big(\{\theta_i^{\top} x_k\}_{k=1}^n;\{y_k\}_{k=1}^n\big),\\
        \theta_i\sim\text{Unif}(\mathbb{S}^{d-1}).
    \end{aligned}
\end{equation}

Here \(n\) is the number of samples, \(m\) represents the number of projections, and \(\hat{I}(\cdot;\cdot)\) is mutual information estimated via KSG estimator~\cite{kraskov2004estimating}. SMI quantifies whether frozen LLM representations retain task-relevant information comparable to fine-tuned models.

\section{Experiments and Results}
\label{sec:results}

\subsection{Experimental Settings}\label{sec:exp}

\begin{table}[htbp]
\scriptsize
\centering
\tabcolsep 0.052in
\begin{tabular}{c|ccccc}
\toprule[1pt]
\textbf{Dataset}    & $\vert \mathbf{\mathcal{E}} \vert$ & \textbf{$\vert \mathbf{\mathcal{R}} \vert$} & \textbf{\# Train} & \textbf{\# Valid} & \textbf{\# Test} \\ \midrule
FB13       & 75,043 & 13     & 316,232  & 5,908  & 23,733  \\
WN11       & 38,696 & 11     & 112,581  & 2,609  & 10,544  \\
FB15K-237N & 13,104  & 93     & 87,282   & 14,082 & 16,452  \\
WN18RR     & 40,943 & 11     & 86,835   & 6,068  & 6,268   \\
UMLS       & 135    & 46     & 5,216    & 1,304  & 1,322   \\ 
YAGO3-10       & 123,182    & 37     & 1,079,040    & 5,000  & 5,000   \\ 
\bottomrule[1pt]
\end{tabular}
\caption{Statistical information of datasets.}
\label{statistics}
\end{table}

\textbf{Datasets.} 
Table~\ref{statistics} summarizes the six benchmark KG datasets: FB13 \cite{socher2013reasoning}, WN11 \cite{socher2013reasoning}, FB15K-237N \cite{lv2022pre}, WN18RR \cite{dettmers2018convolutional}, and UMLS \cite{yao2019kg} for triple classification; YAGO3-10 \cite{dettmers2018convolutional} for relation prediction; and WN18RR for entity prediction. To address incomplete entity descriptions, we generate them from one-hop subgraphs, either by concatenating triples (\textit{Tri}) or further rephrasing via GPT-3.5 (\textit{GPT}). To assess the quality, we also utilize existing entity descriptions (provided by the external authoritative sources) as reference baselines (\textit{Non-Generated}). For negative sampling, we follow standard practice~\cite{yao2019kg}, generating equal numbers of negative samples by corrupting head or tail entities within positive triples.

\textbf{Baselines.} 
We compare with structure-based methods (TransE \cite{bordes2013translating}, DistMult \cite{yang2014embedding}, ComplEx \cite{trouillon2016complex}, RotatE \cite{sun2019rotate}) and additional information-based methods (KG-BERT \cite{yao2019kg}, KG-T5 \cite{saxena2022sequence}, LLaMA-ICL~\cite{zhang2024making}, KG-LLAMA~\cite{yao2025exploring}). Notably, KG-LLAMA parameter-efficient fine-tunes LLaMA for KGC via instruction tuning, relying solely on the LLM's intrinsic capabilities, making it a strong baseline for FLAME. Hybrid methods with external structural embeddings (e.g., KoPA \cite{zhang2024making}) are excluded. FLAME uses a two-layer MLP as the classifier and LLaMA-7B as the backbone language model. All hyperparameters are selected based on validation set performance.

\subsection{Main Results (RQ1)}
\label{sec:rq1}

Table~\ref{general table} shows that FLAME (w/ Non-Generated) and FLAME (w/ GPT), using only 3k training samples (10k for UMLS), achieve comparable performance to KG-LLAMA, which is fine-tuned on the full training set. With equal-sized subsets, they outperform the prior SOTA KG-LLAMA-SAMPLED by 17.15\% and 18.16\% on average.

The ablation results validate our core contributions in §\ref{sec:3.2} and §\ref{sec:3.3}: entity descriptions improve LLM's entity understanding (FLAME (w/ GPT) vs. (w/o description)), and our stimulation method effectively extracts knowledge (FLAME (w/o description) vs. LLaMA variants). UMLS shows an exception where descriptions hurt performance, as its precise medical terminology requires limited room for multiple interpretations. The semantics of entities in the KG have well-aligned with the LLM. Performance on WN11 is slightly lower than KG-LLAMA, likely due to missing knowledge of certain triples during LLM pretraining.

\begin{table}[thb]
\centering
\scriptsize
\begin{tabular}{l|c|ccccc}
\toprule[1pt]
\textbf{Method} & \makebox[0.05in]{\textbf{Samples}}         & \makebox[0.1in]{\textbf{FB13}}   & \makebox[0.1in]{\textbf{WN11}}  & \makebox[0.15in]{\textbf{15K237N}} & \makebox[0.1in]{\textbf{WN18RR}} & \makebox[0.1in]{\textbf{UMLS}}   \\ \midrule
TransE         & ALL   & 0.815  & 0.759 & 0.697     & 0.884      & 0.845 \\
DistMult      & ALL    & 0.862  & 0.871 & 0.587     & 0.851      & 0.864 \\
ComplEx      & ALL   & 0.857  & 0.862 & 0.657      & 0.841     & 0.871 \\ 
RotatE      & ALL   & 0.819  & 0.847 & 0.685      & 0.882     & \underline{0.877} \\ \midrule
KG-BERT     & ALL      & 0.899  & 0.935 & 0.560     & 0.916      & 0.773  \\
KGT5         & ALL     & 0.663  & 0.728 & -          & -      & -      \\
LLAMA-7B       & 0   & 0.699  & 0.661 & 0.573     & 0.458      & 0.658 \\
LLAMA-7B-ICL       & 2 & 0.501   & 0.500   &0.578 &  0.502  &0.538 \\
KG-LLAMA-7B    & ALL   & 0.892  & \textbf{0.955} & \underline{0.748}     & 0.921      & 0.858 \\
KG-LLAMA-SAMPLED   & 3k/10k   & 0.727  & 0.682 & 0.614     & 0.574      & 0.834 \\ \midrule
FLAME (w/o description)    & 3k/10k & 0.855 & 0.872 & 0.674     & 0.858  & \textbf{0.882}  \\
FLAME (w/ Non-Generated)     & 3k/10k & 0.901  & -     & 0.738     & 0.934  & 0.862  \\
FLAME (w/ Tri)  & 3k/10k & 0.842  & 0.755     & 0.671     & 0.779  & 0.782  \\
FLAME (w/ GPT)  & 3k/10k & 0.912   & 0.917 & 0.726      & 0.924  & 0.860  \\ 
FLAME (w/ Non-Generated)   & ALL & \underline{0.920}  & -     & \textbf{0.753}     & \textbf{0.943}  & 0.863  \\
FLAME (w/ GPT)   & ALL & \textbf{0.925}  & \underline{0.937}     & 0.744     & \underline{0.938}  & 0.866  \\
\bottomrule[1pt]
\end{tabular}
\caption{Accuracy of triple classification. `Samples' shows the number of training samples (both positive and negative). The best result per dataset is bolded, and the second-best is underlined.}
\label{general table}
\end{table}

In addition to triple classification, we evaluate FLAME on relation and entity prediction (Table~\ref{combined}). With only 0.6\% of training data, FLAME maintains 97\% of full training set fine-tuning performance in relation prediction. For entity prediction, we obtain the plausibility confidence scores of the triples through the classifier and rank them accordingly. Table~\ref{combined} also displays similar results, showing that FLAME (w/ GPT) outperforms both the FLAME (w/o description) and fine-tuned counterparts.

\begin{table}[htbp]
\scriptsize
\centering
\begin{tabular}{lcc|cc}
\toprule[1pt]
\multirow{2}{*}{\textbf{Method}} & \multicolumn{2}{c|}{\textbf{Relation Prediction}} & \multicolumn{2}{c}{\textbf{Entity Prediction}} \\ 
\cline{2-5}
 & \textbf{Samples} & \textbf{YAGO3-10} &  \textbf{Samples} & \textbf{WN18RR} \\ \midrule
KG-BERT           & ALL   & 0.6816   & ALL   & 0.1102 \\
KGT5                & ALL   & 0.5714    & ALL   & 0.1011 \\
ChatGLM-6B          & 0     & 0.0658  & -     & -         \\
KG-ChatGLM-6B       & ALL   & 0.5662     & ALL   & 0.1613 \\
LLaMA-7B            & 0     & 0.0348    & 0     & 0.0849 \\
LLaMA-13B          & 0     & 0.0040     & 0     & 0.0991 \\
KG-LLaMA-7B        & ALL   & \textbf{0.7028}     & ALL   & 0.2415 \\ \midrule
FLAME (w/o description)      & 6996  & 0.5968  & 10k   & 0.2108 \\
FLAME (w/ GPT)  & 6996 & 0.6824  & 10k  & \underline{0.2433} \\
FLAME (w/ GPT)  & ALL  & \underline{0.7015} & ALL  & \textbf{0.2495} \\ 
\bottomrule[1pt]
\end{tabular}
\caption{Experiment results (Hits@1) for relation prediction and entity prediction.}
\label{combined}
\end{table}

\begin{figure}[htb!]
    \centering
    \includegraphics[width=0.7\linewidth]{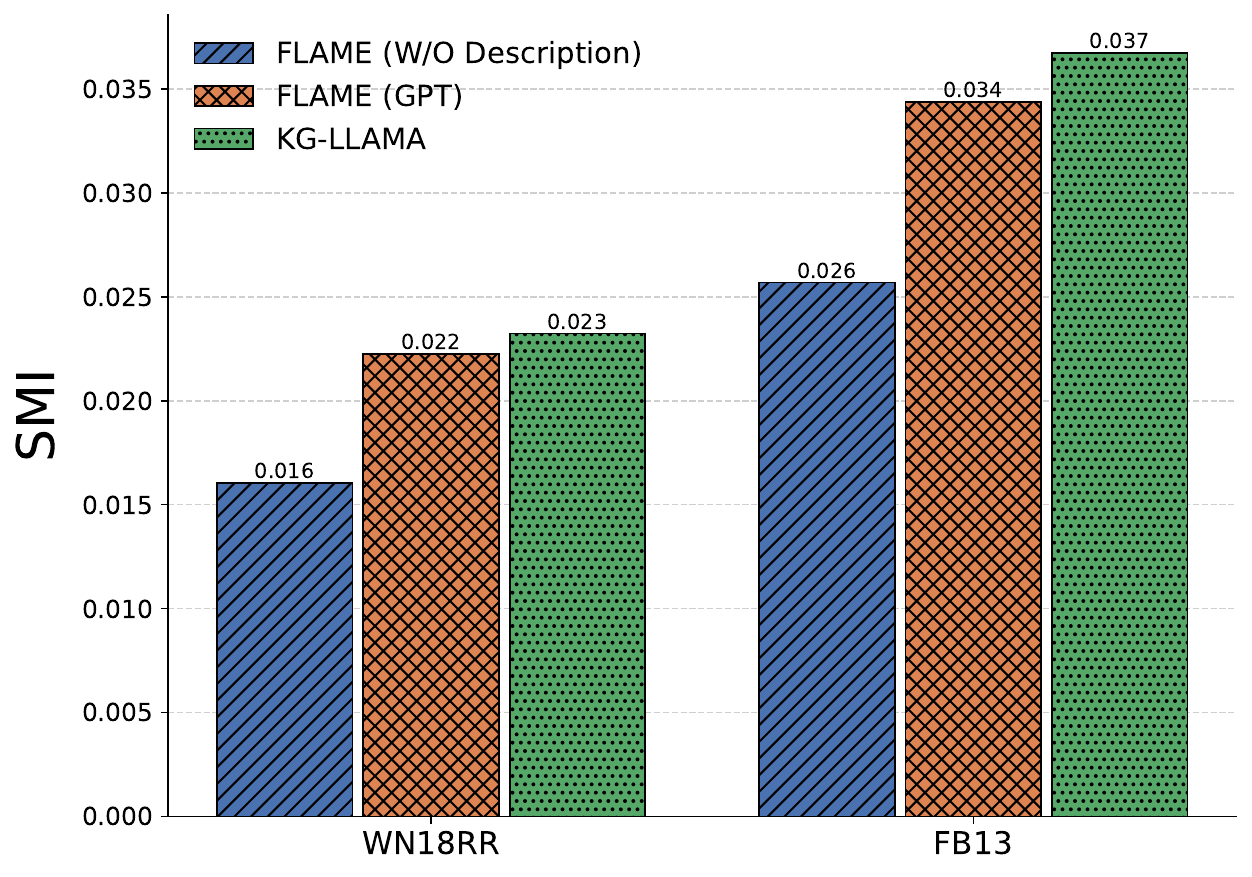}
    \caption{SMI values of intermediate layer representations in the LLaMA model across different methods.}
    \label{fig:smi}
\end{figure}

We quantify task-relevant information in intermediate representations using SMI (Figure~\ref{fig:smi}). Model friendly entity descriptions boost SMI values by 34.1\%, validating our generator's effectiveness in aligning LLM's internal semantic comprehension with KGs' entities. Moreover, these representations reach fine-tuned-level SMI values, indicating that fine-tuning primarily aligns representations rather than injecting knowledge from the KG training set. This information-theoretic characterization of LLM internal representations validates the feasibility and effectiveness of leveraging frozen LLMs for efficient and accurate KGC.

Collectively, these results show that non-fine-tuned LLMs with our method can achieve performance comparable to fully fine-tuned models, effectively answering \textbf{RQ1}.

\subsection{Ablation Study (RQ2)}\label{sec:rq2}

Table~\ref{tb:model cls} shows that FLAME consistently achieves performance comparable to fine-tuned models across diverse LLM architectures, demonstrating strong generalizability. Entity descriptions yield consistent gains of 4.5\% (w/ GPT) and 6.2\% (w/ Non-Generated) over baseline, validating our semantic alignment mechanism's architecture-agnostic effectiveness.

\begin{table}[htbp]
\scriptsize
\centering
\begin{tabular}{c|ccc}
\toprule[1pt]
\textbf{Model}                    & \textbf{Method}    & \textbf{FB13} & \textbf{WN11} \\ \midrule
\multirow{4}{*}{\textbf{Mistral}}   & KG-MISTRAL-7B           & 0.891         & 0.962         \\
  & FLAME (w/o description)          & 0.846         & 0.901         \\
  & FLAME (w/ Non-Generated)        & 0.880         & -             \\
  & FLAME (w/ GPT)   & 0.893         & 0.925         \\ \midrule
\multirow{4}{*}{\textbf{Gemma}}   & KG-GEMMA-7B           & 0.897             & 0.944             \\
  & FLAME (w/o description)          & 0.806         & 0.893         \\
  & FLAME (w/ Non-Generated)         & 0.891         & -             \\
  & FLAME (w/ GPT)   & 0.867         & 0.917         \\ 
  \midrule
\multirow{4}{*}{\textbf{Qwen2.5}}   & KG-Qwen-7B           & 0.913             & 0.961             \\
  & FLAME (w/o description)         & 0.864         & 0.908         \\
  & FLAME (w/ Non-Generated)        & 0.904         & -             \\
  & FLAME (w/ GPT)   & 0.916         & 0.935         \\ 
\bottomrule[1pt]
\end{tabular}

\caption{KGC performance on FB13 and WN11 using Mistral~\cite{jiang2023mistral}, Gemma~\cite{team2024gemma}, and Qwen2.5~\cite{qwen2025qwen25technicalreport} as base models}
\label{tb:model cls}
\end{table}

We also compare classifiers (MLP, SVM, Logistic Regression) in Table~\ref{cls table}. MLP consistently performs best, with Logistic Regression close behind, indicating that hidden states already encode discriminative features for KGC, enabling effective classification with simple models.

\begin{table}[htbp]
\scriptsize
\centering
\tabcolsep 0.1in
\begin{tabular}{c|ccc}
\toprule[1pt]
\textbf{Method}       & \textbf{FB13}   & \textbf{WN11}   & \textbf{FB15K-237N} \\ \midrule
\textbf{SVM} & 0.821 & \underline{0.866} & 0.594      \\
\textbf{Logistic-Regression}  & \underline{0.839}   & 0.860 & \underline{0.662}      \\
\textbf{MLP} & \textbf{0.855} & \textbf{0.872}  & \textbf{0.674}     \\ \bottomrule[1pt]
\end{tabular}
\caption{Performance comparison of FLAME (w/o description) across different classifiers.}
\label{cls table}
\end{table}

Our method performs robustly across different LLM backbones and classifiers, confirming its generalizability and addressing \textbf{RQ2}.

\subsection{Efficiency Study (RQ3)}
\label{sec:rq3}

\begin{figure}[htb]
    \centering
    \includegraphics[width=0.75\linewidth]{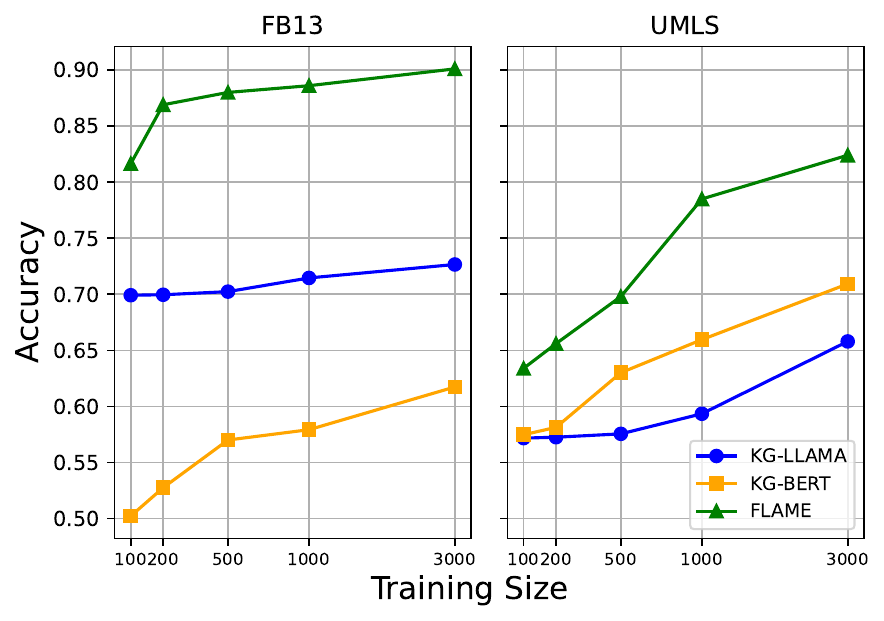}

    \caption{Performance of KG-LLAMA, KG-BERT, and FLAME (w/ Non-Generated) on triple classification with varying training sizes.}
    \label{fig:ts}
    
\end{figure}

Fig.~\ref{fig:ts} shows FLAME outperforms KG-LLAMA and KG-BERT with merely 100 samples versus their 3000 samples on FB13. With 500 samples (0.079\% of FB13), FLAME achieves 97.2\% of full performance versus 78.7\% and 63.1\% for KG-LLAMA and KG-BERT. On UMLS, FLAME with 500 samples matches baselines' 3000-sample performance; with 9.58\% of training data, it reaches 90.6\% of full performance. These results validate FLAME's ability to leverage LLM internal representations for data-efficient KGC.

Table~\ref{tab:efficiency} shows FLAME reduces GPU memory by $188\times$ during training and achieves $26.11\times$ speedup compared to parameter-efficient fine-tuned KG-LLAMA (LoRA~\cite{hu2022lora} combined with 8-bit quantization). The training overhead is limited to the classifier and independent of LLM size, making our approach increasingly advantageous for larger models (e.g., 70B) where fine-tuning is prohibitively expensive.

\begin{table}[htbp]
\scriptsize
\setlength\tabcolsep{0.5pt}
\begin{tabular}{c|ccc}
\toprule[1pt]
\multicolumn{1}{c|}{\textbf{Method}} & \textbf{Procedure}                                                    & \textbf{GPU Memory}                                                                    & \textbf{Time}                                                      \\ \midrule
\multirow{2}{*}{\textbf{KG-LLAMA}}   & \begin{tabular}[c]{@{}c@{}}Training\\ (LLM:1 epoch)\end{tabular}      & \begin{tabular}[c]{@{}c@{}}14.68G\\ {\fontsize{5.5pt}{6.5pt}\selectfont (LLM Parameters+LoRA+Gradient etc)} \end{tabular}    & \begin{tabular}[c]{@{}c@{}}83h\\ {\fontsize{5.5pt}{6.5pt}\selectfont (Forward+Backward)}\end{tabular}   \\ \cmidrule{2-4} 
                                     & \begin{tabular}[c]{@{}c@{}}Inference\\ (LLM:Generation)\end{tabular}  & \begin{tabular}[c]{@{}c@{}}12.94G\\ {\fontsize{5.5pt}{6.5pt}\selectfont (LLM Parameters+LoRA)}\end{tabular}                 & \begin{tabular}[c]{@{}c@{}}2h50min\\ {\fontsize{5.5pt}{6.5pt}\selectfont (Forward)}\end{tabular}       \\ \midrule
\multirow{2}{*}{\makecell{\textbf{FLAME}}}  & \begin{tabular}[c]{@{}c@{}}Training\\ (MLP:10 epochs)\end{tabular}     & \begin{tabular}[c]{@{}c@{}}0.078G\\ {\fontsize{5.5pt}{6.5pt}\selectfont (MLP Parameters+Gradient etc)}\end{tabular}         & \begin{tabular}[c]{@{}c@{}}33min\\ {\fontsize{5.5pt}{6.5pt}\selectfont (Forward+Backward)}\end{tabular} \\ \cmidrule{2-4} 
                                     & \begin{tabular}[c]{@{}c@{}}Inference\\ (LLM:Probing+MLP)\end{tabular} & \begin{tabular}[c]{@{}c@{}}12.82G+0.018G\\ {\fontsize{5.5pt}{6.5pt}\selectfont (LLM Parameters+MLP Parameters)}\end{tabular} & \begin{tabular}[c]{@{}c@{}}2h44min+15s\\ {\fontsize{5.5pt}{6.5pt}\selectfont (Probing+MLP Forward)}\end{tabular}    \\ \bottomrule
\end{tabular}
\caption{Efficiency comparison on the WN11 full dataset. The `Procedure' column represents several key steps of these methods.}
\label{tab:efficiency}
\end{table}

Our method is highly data- and computation-efficient, achieving strong KGC performance with minimal training data and resources, thus answering \textbf{RQ3}.

\section{Related Work}\label{sec:related}

KGs are inherently incomplete, necessitating effective approaches for knowledge graph completion \cite{hogan2021knowledge}. Existing KGC methods fall into two paradigms: structure-based and additional information-based methods  \cite{shen2022comprehensive}. Structure-based methods exploit graph topology through embedding-based scoring functions, including TransE  \cite{bordes2013translating}, DistMult  \cite{yang2014embedding}, ComplEx  \cite{trouillon2016complex}, ConvE  \cite{dettmers2018convolutional} and RotatE  \cite{sun2019rotate}. Additional information-based methods incorporate supplementary data to enrich KGs. These include KG-BERT  \cite{yao2019kg} and LASS  \cite{shen2022joint}, which reformulate KGC as sequence classification using encoder-only models, while KG-S2S \cite{chen2022knowledge} and KGT5 \cite{saxena2022sequence} adopt generative paradigms with encoder-decoder architectures. Recent advances leverage decoder-only LLMs for KGC. KoPA \cite{zhang2024making} combines structural embeddings with LLMs for joint structural-contextual modeling. More recently, KG-LLAMA \cite{yao2025exploring} frames KGC as an instruction-based question-answering task, using parameter-efficient fine-tuning. Unlike hybrid approaches that combine LLMs with external structural embeddings, KG-LLAMA solely relies on the intrinsic capabilities of LLaMA, aligning the core motivation of our proposed framework.

\section{Conclusion}\label{sec:conclusion}
We propose FLAME, a novel framework that leverages frozen LLM latent representations for efficient KGC. By integrating subgraph-based entity descriptions, FLAME bridges the LLM-KG semantic gap and achieves fine-tuning-level performance without parameter updates. It also achieves strong data efficiency while substantially reducing GPU memory usage and training time. FLAME effectively mitigates the efficiency-performance trade-off in KGC, offering a principled alternative to fine-tuning.

\bibliographystyle{IEEEbib}
\bibliography{refs}

\end{document}